\documentclass[10pt, a4paper]{article}
\usepackage{lrec2000}
\usepackage{epsfig}

\title{Using the Annotated Bibliography as a Resource for Indicative Summarization}

\name{Min-Yen Kan, Judith L. Klavans, and Kathleen R. McKeown} 

\address{Department of Computer Science \\
	 Columbia University \\
	 New York, New York  10027 USA \\
	 {\tt \{min,klavans,kathy\}@cs.columbia.edu}
}

\abstract{We report on a language resource consisting of 2000
annotated bibliography entries, which is being analyzed as part of our
research on indicative document summarization.  We show how annotated
bibliographies cover certain aspects of summarization that have not
been well-covered by other summary corpora, and motivate why they
constitute an important form to study for information retrieval.  We
detail our methodology for collecting the corpus, and overview our
document feature markup that we introduced to facilitate summary
analysis.  We present the characteristics of the corpus, methods of
collection, and show its use in finding the distribution of types of
information included in indicative summaries and their relative
ordering within the summaries.}

\begin{document}

\maketitleabstract



\section{Introduction}

Automatic text summarization has largely been synonymous with
domain-independent, sentence extraction techniques (for an overview,
see Paice~\shortcite{Paice90}).  These approaches have used a battery
of indicators such as cue phrases, term frequency, and sentence
position to choose sentences to extract and form into a summary.  An
alternative approach is to collect sample summaries and apply machine
learning techniques to identify what types of information are included
in a summary, and identify their stylistic, grammatical, and lexical
choice characteristics and to generate or regenerate a summary based
on these characteristics.  In this paper, we examine the first step
towards this goal: the collection of an appropriate summary corpus.
We focus on annotated bibliography entries, because they are written
without reliance on sentence extraction.  Futhermore, these entries
contain both informative (i.e., details and topics of the resource) as
well as indicative (e.g., metadata such as author or purpose)
information.  We believe that summary texts similar in form to
annotated bibliography entries, such as the one shown in Figure
\ref{f:sampleABE}, can better serve users and replace standard $n$-top
sentence or query word in context summaries commonly found in current
generation search engines.

Our corpus of summaries consists of 2000 annotated bibliography
entries collected from various Internet websites using search engines.
We first review aspects and dimensions of text summaries, and detail
reasons for collecting a corpus of annotated bibliography entries.  We
follow with details on the collection methodology and a description of
our annotation of the entries.  We conclude with some current
applications of the corpus to automatic text summarization research.

\begin{figure}[h]
\centering
\small
\framebox[3.25in]{
\parbox{3in}{

{\bf Maxwell, S. E., Delaney, H. D., \& O'Callaghan,
M. F. (1993). Analysis of covariance. In L. K. Edwards (Ed.), Applied
analysis of ... } \\

\begin{tabular}{p{.1in}p{2.7in}}
& This paper gives a brief history of ANCOVA, and then
discusses ANCOVA in the context of the general linear model. The
authors then provide a numerical example, and discuss the assumptions
of ANCOVA. Then four advanced topics are covered: ... This paper is
quite theoretical and complex, but contains no matrix algebra. \\
\end{tabular}
}}

\caption{Sample excerpt from an annotated bibliography entry.}
\label{f:sampleABE}
\end{figure}

\section{Dimensions of summarization}

With the current widespread language resources that are available on
the web, constructing a large corpus of document summaries is becoming
easier.  However, document summaries have many different aspects and
purposes (Mani and Maybury~\shortcite{Mani&Maybury99}, introduction),
and thus it is important to clarify which aspects of summarization a
collection covers.  We briefly examine several different dimensions of
summaries.

\begin{itemize}
\item {\bf Extract versus Abstract} - Summaries that are constructed
by extracting important passages, sentences or phrases from the source
document are considered {\it extracts}.  In contrast, an abstract may
or may not contain words in common with the document.  Authors using
abstractive techniques are not as constrained as those using
extractive ones, and can summarize a wider range of materials
effectively (e.g., narratives) and often with smaller amounts of text.
\item {\bf Informative versus Indicative} - Informative summaries
attempt to include all important points of the document in the
summary.  Examples include book reports or scientific abstracts of
technical articles.  Indicative summaries hint at the topics of the
document, and do not serve as any type of surrogate for the source
document.  From an information retrieval perspective, we can think of
the indicative summary as text that helps a user to decide whether
they should consider retreiving the full text of the source document.
Examples of indicative summaries include annotated bibliography
entries and library card catalog entries.
\item {\bf Generic versus Query-based} - Summaries that treat all
topics of a source document with equal weight are generic summaries,
whereas a query-based summary gives particular attention to a specific
facet of the document.  While library card catalog entries are generic
summaries, annotated bibliography entries that are part of a themed
collection (e.g., ``Books about Medieval Arms and Armor'') are often
biased towards the collection's topic, and may highlight or only
mention information relating to its theme.
\item {\bf Single Document versus Multidocument} - Multidocument
summaries typically summarize a set of documents that are related in
some fashion.  Current multidocument summary techniques have focused
on articles provided by different sources, or which are updates of
previous articles on an event \cite{Radev&McKeown98}.
\end{itemize}

\section{Related work in summary corpora}

With these dimensions of text summarization in mind, we can discuss
different existing summary corpora, and show how they relate to these
particular dimensions.  This is shown in Table \ref{t:corporaTypes}.

\subsection{News summaries}

The Document Understanding Conference (DUC) was first held in 2001,
sponsored by the National Institute of Science and Technology (NIST)
\cite{DUC01}.  It is a competition in the ``bake-off'' style which
pits systems against each other in summarizing the same set of input
documents.  For the first DUC competition, training corpora of sample
input documents and sample summaries were provided by NIST in
consultation with the research community.  Both single document and
multidocument generic summaries were made available to groups to train
15 different summarization systems.  The DUC summary corpus was
constructed by both extractive and abstractive techniques, and tend to
be informative rather than indicative.

Jing and McKeown \shortcite{Jing&McKeown99} also have made use of
source document and target summary relation, in their use of Hidden
Markov Models for summarization.  Their ``cut and paste'' method was
demonstrated on the Ziff-Davis summary corpus of computer
peripheral review articles.  The Ziff-Davis summary corpus is a
single document corpus that is generic and mostly extract-based.

\subsection{Scientific summaries}

There have been a number of studies using abstracts of scientific
articles as a target summary.  Kupiec, Pedersen and Chen
\shortcite{KupiecEtAl95}'s work is an instance of this, where they use
188 {\it Engineering Information} summaries that are mostly indicative
in nature.  Abstracts tend to summarize the document's topics well but
do not include much use of metadata, which is of interest to our
study, further explained in Section \ref{s:abe}

\subsection{Snippets}

Snippets \cite{Amitay00}, are short, textual descriptions that authors
of web pages provide to give an indicative description to a
hyperlinked document.  These snippets are often very short, as in the
case of the descriptions connected to Yahoo! or Open Directory Project
(ODP) category pages.  Amitay describes strategies for locating and
extracting snippets from various types of web pages, and applies
machine learning to rank different snippet description of the same
document for fitness as a document summary.

This solution only works for resources that have existing snippets.
Newly-authored documents (of interest to people trying to keep
current) cannot benefit from past snippets, since they refer to
different resources.  Amitay's work lays the foundation for building
the tools to collect such a snippet corpus, but unfortunately does not
provide a publically available tool nor corpus.

\subsection{Card catalog summaries}
\label{ss:cardCatalogSummaries}

Library card catalog entries in the physical library (and their
electronic, machine-readable record conterparts in the automated
library) also provide indicative summaries of resources.  Our
preliminary study \cite{KanEtAl01} examined these resources to get a
first-round approximation of the contents of indicative summaries.
Library catalog entries consist of structured fields, of which a
summary is an optional field.  These summary fields are often provided
by third-party vendors who may not be aware of the other fields
present in the catalog.  In our local online catalog, other types of
information (such as notes, or book jacket texts, or book reviews)
were often substituted for summaries.

\begin{table*}[hbt]
\centering
\begin{tabular}{|l|l|l|l|l|l|l|}
\hline
Corpus & Extract vs. & Indicative vs. & Generic vs. & Single vs.    & Uses      & Corpus vs. \\
       & Abstract    & Informative    & Query-based & Multidocument & Metadata? & Algorithm \\
\hline
\hline
DUC & Both & Informative & Generic & Both & Yes & Corpus \\
Ziff-Davis & Extract & Informative & Generic & Single & No & Corpus \\
Scientific Abstracts & Abstract & Indicative & Generic & Single & No & Corpus \\
Snippets & Abstract & Indicative & Both & Single & Yes & Algorithm \\
Card Catalog Entries & Abstract & Indicative & Generic & Single & Yes & Corpus \\
Annotated Bibliography & Abstract & Both & Both & Mostly Single & Yes & Corpus \\
\hline
\end{tabular}
\caption{Sample summary corpora types mentioned in this paper.}
\label{t:corporaTypes}
\end{table*}

\section{Annotated bibliography entries}
\label{s:abe}

Broadly speaking, our research focuses on how automatic text
summarization techniques can be applied to understanding search engine
results.  Our goal is not to analyze what makes one summary better
than another, but to learn how to generate a suitable summary of a
resource based on machine learning over a compiled corpus.  A
``suitable'' annotation can span many different dimensions, but in our
case mainly concerns space/length limitations.  Current standard
technology presents search results as a ranklist of 10 or 20 document
``hits'', accompanied by short extract summaries.  An alternative
approach is to present the documents with more meaningful summaries
that explicitly assist the user in choosing a document to examine or
in deciding that none of the retrieved documents are useful.


To fulfill this purpose, query-based indicative summaries constructed
by abstractive techniques are most relevant.  We believe abstracts are
more powerful than extracts because they have the capability to yield
more concise and accurate summaries.  Similarly, indicative
summarization is an equally important facet, as it provides summaries
tailored to our information retrieval application, in which source
documents are readily available.  For these reasons, both the DUC and
Ziff-Davis corpora are not well suited to our study.  Scientific
abstracts and library card catalog summaries are largely generic and
thus do not give us an opportunity to study query-based summarization.
The study of snippets most closely aligns with the purpose of our
study, but a compiled corpus of snippets is not publically available,
neither is a tool for locating them.

Instead, we examined a different class of summary texts, the annotated
bibliography entry.  Annotated bibliographies are created mostly by
abstractive methods and include both indicative and informative forms.
An annotated bibliography entry is a summary of a book or other
resource that annotates a resource with a description of the text, as
shown in Figure \ref{f:sampleABE}.

From our empirical observations of both annotated bibliography
entries, snippets and library card catalog entries, bibliography
entries have some unique features that make them attractive and
challenging to process.  Bibliography entries often:

\begin{itemize}
\item are lengthier than both card catalog summaries and snippets.
They often exhibit more variation of sentence structure and lexical
choice.  This makes the subsequent analyses rich and allows
(re)generation based on these analyses to construct more varied and
interesting text.
\item are organized around a theme, making them ideal standard for
``query-based'' summaries.  Bibliography entries also have more explicit
comparison of one resource versus another, which can help a user
determine whether which document to choose for a particular purpose.
\item have prefacing text that overviews the documents in the
bibliography.  This preface text is a good model for summarizing a set
of related items (e.g., different books on arms and armor or different
earthquakes reports in 1992).  This is in contrast to multidocument
summaries that summarize articles with mostly overlapping information
(news reports on a single event and updates to the event).
\item are rich in meta-information document features ---they often
mention edition, title, author and purpose.  These document features
are not always present in or inferrable from the body text of a source
document.  Our previous study of library card catalog entries showed
that these document features are well represented (and thus
important).
\end{itemize}

The construction of annotated bibliographies is a well-established
field in information science studies.  Thus, the form has many
descriptive guidelines that we examined that validate the above
observations.  Writing guides such as
\cite{Rees70,EngleEtAl98,Lester01,AACC98,Williams02} indicate specific
types of information that should be included in annotated
bibliographies; and are synopsized in Table
\ref{t:referenceABEfeatures}.

\begin{table}[htb]
\centering
\scriptsize
\begin{tabular}{|l|c|c|c|c|c|}
\hline
		& Ree70& EBC98 	& Les01 & AACC98 & Wil02\\
\hline
\hline
Accuracy/Currency& 	& X	&	& X	& \\
Audience	&  	& X	& X	& X	& X \\
Authority 	&	& X 	&	& X	& X \\
Cross-resource Comparison& 	& X	&	&	& \\
Contents	&	&	& X	& 	& X \\
Coverage	& X	&	&	& X	& \\
Defects/Weakness&	&	& X	&	& X \\
Navigation	&	&	& 	& X	& \\
Purpose		& X	&	& X	& X	& \\
Quality 	& 	& X	&	& 	& \\
Relevance 	& 	& X	&	& 	& X \\
Subjective Assessment&	& X	&	&	& X \\
Special Features&	&	& X	& X	& \\
\hline
\end{tabular}
\caption{Prescribed features of annotated bibliographies from several
sources}
\label{t:referenceABEfeatures}
\end{table}

These resources are all guidelines for the content of annotated
bibliography entries.  The guidelines are prescriptive, and thus, it
is important to validate them by examining actual annotated
bibliographies to see whether a) the guidelines on content are
followed, and b) to establish the content's ordering and grammatical
structure.

\section{Annotated bibliography language resource}

Our language resource of annotated bibliography entries was designed
to ease the collection of the corpus as well as to make many features
available for subsequent analysis for summarization and related
natural language applications.

\subsection{Collection methodology}

The collection of the bibliography entries was done by spidering
search result pages from two search engines (AltaVista and Google) for
the keywords ``annotated bibliography''.  The collection was compiled in
September 2001 and software filters were written to parse and retrieve
the contained URLs from each site (200 from AltaVista and an additional
1000 from Google).  By our estimates, roughly 60\% of the pages that
were gathered had errors in retrieval (e.g., were stale URLs), were
duplicate entries, or did not contain bibliographic entries.  This
leaves an approximate 500 pages with actual bibliographic entries to
draw from.

An examination of the materials in these remaining documents revealed
that most pages organized around a specific purpose, and varied
greatly in collection size.  Most common were large collections of 20
to 100 entries and introductory pages to even larger collections (over
1000 entries).  Pages that only annotated a few items were much less
common; we suspect that this is due to the inherent bias of the search
engine ranking metric to rank sites that are more prominent (which we
believe is highly correlated with larger collections).  The smaller
collections were often a part of a larger website or were the last
section of a larger webpage on the topic of interest.  With this
structure in mind, we decided to take at most 50 entries from each
source document to ensure that we covered a breadth of annotated
bibliography entry sources in collecting the final corpus.  We
examined the documents in order of their appearance on the AltaVista
hitlist, and as a result, only a total of 64 documents from the
AltaVista spidered collection were used to create the 2000-entry
corpus.  If all of the bibliographic entries were extracted from the
documents, it would easily exceed 20,000 entries in size (as many of
the collections had many more than 50 entries).  Documents spidered
from Google have so far not processed and added to the bibliography
collection; we plan to include the processing of these documents and
other sources as future project time allows.

\subsection{Encoding the XML bibliographic entry corpus}

Bibliography entries from the 64 spidered pages were then manually
cut-and-pasted into the corpus collection web interface.  This was
both to ensure that the entries were being correctly delimited, and to
add fields to each entry that may assist in future analysis and serve
as a gold standard for future machine learning tasks.  The corpus is
encoded in XML and includes the following fields in addition to the
bibliographic entry itself.

\begin{figure}[ht]
\centering
\epsfig{file=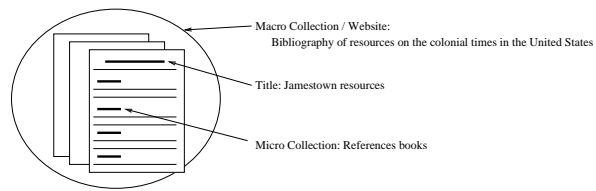,width=3.25in}
\caption{Relation of micro and macro collection attributes}
\label{f:microMacro}
\end{figure}

\begin{itemize}
\item Subject: the subject or theme of the annotated bibliography page.

\item Domain: annotated to aid analysis of differentiation of features
that are domain-independent from ones that are domain-dependent.  We
encode the domain rather coarsely (e.g., all of {\tt medicine} as a
single domain) and in an ad-hoc manner without the assistance of an
ontology.  Finer granularity is provided by the above {\tt subject}
field.

\item Micro Collection (optional): the internal division in the
bibliography page that the entry is a part of (e.g., ``reference
books'' section of a bibliography on the colonial times in Jamestown).

\item Macro Collection (optional): the division that the physical
bibliography page represents in the set of related bibliography pages
(e.g., ``all colonies in colonial times in the U.S.'' with respect to
the last example).  The macro collection field is used when the
bibliography physical page relates itself to other physical pages.  In
our observations, only very large collections exhibit both micro and
macro collection attributes.  Figure \ref{f:microMacro} illustrates
the relation of these two attributes.

\item Offset: the position of the entry on the page.

\item Before Context: text before the body of the annotated entry
itself.  This often contains cataloging and bibliographic information,
such as the title, author, and call number\footnote{Currently, this is
saved as an unstructured text field.  It would be best to parse these
entries into structured fields but our focus is on the text and
content of the entries themselves, and not these auxiliary fields.}.

\item After Context (optional): text that is distinctly marked off as
coming after the body of the annotated entry.  Used sometimes to mark
publisher information, web URLs and pointers to other resources.
Information that typically is contained in this field in one document
may simply be appended to the end of the bibliographic entry in other
documents; this distinction may be more of a stylistic one.

\item URL: the location of the source document where the entry was
drawn from.
\end{itemize}

To facilitate our local analysis of the corpus, all of the
bibliographic entries have also been parsed with a probabilistic
dependency parser \cite{Collins96}.  These parsed entries are also
included in the XML corpus, as a separate XML field attached to each
entry (the {\tt parsedEntry} field).  Figure \ref{f:sampleABEinCorpus}
shows a sample entry after it has been parsed into our XML format.



\begin{figure}[h]
\scriptsize
\begin{verbatim}
<bibEntry id="id26" title="Analysis of covariance" 
  url="http://www.math.yorku.ca/SCS/biblio.html" 
  type="paper" domain="statistics"
  microCollection="Analysis of Covariance"
  offset="4">
  <beforeContext>
    Maxwell, S. E., Delaney, H. D., & O'Callaghan, 
    M. F. (1993). Analysis of ...
  </beforeContext>
  <entry><OVERVIEW>This <MEDIATYPES>paper</MEDIATYPES> 
    gives a brief history of ANCOVA, and then discusses 
    ANCOVA in ... contains no matrix 
    algebra.</DIFFICULTY>
  </entry>
  <parsedEntry> 
    PROB 14659 -112.252 0 TOP -112.252 S -105.049 NP-A 
    -8.12201 NPB -7.82967 DT 0 This NN 0 paper ... 
  </parsedEntry>
</bibEntry>
\end{verbatim}
\caption{Portion of the annotated bibliographic entry from Figure
\ref{f:sampleABE}, represented as structured fields in our XML corpus.}
\label{f:sampleABEinCorpus}
\end{figure}
\subsection{Semantic annotation of document features}

To perform a detailed study of what information is normally present in
annotated bibliographic entries, we needed to inventory the different
document features (types of information) used in the entries.  We
re-used our original 14 document features used in our earlier work on
library card catalog entries (as mentioned in Section 3.4) and further
enriched the feature set to include additional tags that better
represent the range of information we found in the annotated
bibliography entries.  We also took into account annotated
bibliographic guidelines, as mentioned in Section \ref{s:abe} We
randomly picked 100 of the 2000 entries to annotate using this scheme.
Table \ref{t:documentFeatures} shows the expanded, 24 document feature
set used in the markup.

\begin{table*}[htb]
\centering
\begin{tabular}{|l|r|r|}
\hline
Document Features & \# tag occurrences & \% entries possessing tag \\
          & (tag frequency) & (document frequency) \\
\hline
\hline
\multicolumn{3}{|l|}{{\bf Topicality document features} - features based on contents of the body text}\\
\hline
\vspace{-.2cm}
Detail & 139 & 47\% \\
{\scriptsize Quotations, extracted sentences, parts of a chronology, conclusions} & & \\ 
\vspace{-.2cm}
Overview & 72 & 64\% \\
{\scriptsize (Generalized description of the entire resource, ``This book is about Louisa Alcott's life.'')} & & \\ 
\vspace{-.2cm}
Topic & 34 & 28\% \\
{\scriptsize (High-level {\bf list} of topics, e.g., ``Topics include symptoms, ...'')} & & \\ 
\hline
\hline
\multicolumn{3}{|l|}{{\bf Metadata and document-derivable features} - features that are domain- and genre-independent} \\
\hline
\vspace{-.2cm}
Media Type & 55 & 48\% \\
{\scriptsize (e.g. ``This book ...'', ``A weblet ...'', ``Spans 2 CDROMs'')} & & \\ 
Author / Editor* & 43 & 27\% \\
\vspace{-.2cm}
Content Types & 41 & 29\% \\
{\scriptsize (e.g. ``figures and tables'')} & & \\ 
\vspace{-.2cm}
Subjective Assessment* & 36 & 24\% \\
{\scriptsize (e.g. ``highly recommended'')} & & \\ 
Authority / Authoritativeness* & 26 & 20\% \\
\vspace{-.2cm}
Background / Source* & 21 & 16\% \\
{\scriptsize (e.g. ``based on a report'')} & & \\ 
\vspace{-.2cm}
Navigation / Internal Structure* & 16 & 11\% \\
{\scriptsize (e.g. ``is organized into three parts'')} & & \\ 
Collection Size* & 13 & 10\% \\
Purpose* & 13 & 10\% \\
\vspace{-.2cm}
Audience* & 12 & 12\% \\
{\scriptsize (e.g. ``for adult readers'')} & & \\
\vspace{-.2cm}
Contributor* & 12 & 12\% \\
{\scriptsize Name of the author of the annotated entry} & & \\
\vspace{-.2cm}
Cross-resource Comparison* & 10 & 9\% \\
{\scriptsize (e.g., ``similar to the other articles''} & & \\
Size/Length & 9 & 7\% \\
\vspace{-.2cm}
Style* & 8 & 6\% \\
{\scriptsize (e.g., ``in verse rhythm'', ``showcased in soft watercolors'')}& & \\
\vspace{-.2cm}
Query Relevance* & 4 & 3\% \\
{\scriptsize (text relevant to the theme of the annotated bibliography collection)} & & \\
Readability* & 4 & 4\% \\
\vspace{-.2cm}
Difficulty* & 4 & 4\% \\
{\scriptsize (e.g., ``requires no matrix algebra'')}& & \\
Edition / Publication* & 3 & 3\% \\
Language & 2 & 2\% \\
Copyright* & 2 & 1\% \\
Award* & 2 & 1\% \\
\hline
\end{tabular}
\caption{Distribution of the document features in the 100 entry annotated
portion of the corpus.  Starred entries denote metadata fields.}
\label{t:documentFeatures}
\end{table*}

\section{Corpus attributes}


Table \ref{t:documentFeatures} also lists distributional features of
the tagged document features in the 100 annotated entries.  The first
column shows the number of times that the annotated feature was used
to mark information in the entries.  The second column gives the
precentage of documents that have an instance of the feature in
question.  Features were marked at the sentence level or on smaller
units.  The columns are highly correlated, and show that multiple
occurrences of the same tag within an entry happen quite frequently.

We divided the features into topically related and unrelated features.
We distinguish between three different topically related features.
{\it Overview} sentences usually begin the annotated bibliography
entry and include a high level overview of the content of the
resource.  They appear in a majority of annotated bibliography entries
and generally are limited to a single sentence.  {\it Topic} features
give a list of topics treated by the source, as an itemized or
comma-delimited list.  {\it Detail} sentences represent all other
general item-specific sentences.  In our observations across the 100
entries that we annotated, these sentences were the most variable.
Short entries tended not to have any {\it detail} sentences, but as we
examined entries of longer length, mostly {\it detail}s were being
added.

The data validates both prescriptive guidelines and our earlier work
in showing that metadata fields (marked with stars in Table
\ref{t:documentFeatures}) are important for summaries.  {\it Audience}
information, recommended by four of the five prescriptive guidelines,
were shown to appear 12\% of the time.  Other metadata fields, such as
{\it purpose}, {\it navigation/internal structure}, {\it subjective
assessment}, and {\it readability} also play important roles.

A noticeable difference between our earlier work on card catalog
entries is that the {\it title} field does not appear in any of the
annotated bibliography entries.  We surmise this is because its
mention would be redundant, as the title is always given as text in
the {\tt beforeContext} XML field.  However, this is not true of {\it
author} information, as the document feature is often used to present
the credentials of the author.  In contrast, library card catalog
entries did exhibit the {\it title} field quite often.  We feel that
this is because card catalog summaries were often book jacket or other
related standalone texts that may not have easy access to the
bibliographic information.

Table \ref{t:featureDistribution} shows how the distribution of the 24
document features varies with length and indicates where the features
occur within the summary.  The numbers between 0 and 1 in paratheses
indicates how close the average instance of the document feature is to
the beginning (0) of the summary entry or to the end (1).  Middle
range numbers (e.g., .50) often indicate that the field occurred
widely across different positions in the entries, especially when the
feature frequency is high.  Entries tended to include 2 to 6 document
features, and long bibliography entries were fairly rare (entries with
13 or more document feature instances represent only 6\% of the
annotated corpus).  Normal entries containing 2 to 6 document features
correspond to 2 to 4 sentence- or phrase-length entries.

Examining the ordering data, it is quite apparent that some of the
fields naturally occur before or after others.  Overview sentences
generally comes very early in the bibliography entry, and information
on who wrote the entry (the contributor) usually comes very late.
Subjective assessment or critique of a resource usually comes after an
explanation of the resource, thus comes later in the summary.
Ordering among the features is quite variable, but it is obvious that
many of features either tend to occur earlier (e.g., bibliographic
information) or later (e.g., subjective assessment or complicated
types of metadata) with topical information filling in the space
between.

\begin{table*}[htb]
\centering
\tiny
\begin{tabular}{|l|r|r|r|r|r|r|r|r|r|r|r|r|r|r|r|r|}
\hline
Feature & \multicolumn{16}{|c|}{Number of tags in entry} \\
\hline
Entry Length & 1 & 2 & 3 & 4 & 5 & 6 & 7 & 8 & 9 & 10 & 11 & 13 & 14 & 15 & 18 & 20 \\
\hline
\# of Entries of Indicated Length & (4) & (10) & (14) & (16) & (16) & (9) & (5) & (7) & (3) & (5) & (5) & (1) & (2) & (1) & (1) & (1) \\
\hline
\hline
Detail 				&        &        & 8 (.56)&14 (.69)&21 (.64)&18 (.66)& 9 (.50)&13 (.62)& 4 (.50)& 7 (.52)&12 (.58)&        & 6 (.63)& 6 (.48)&16 (.56)& 5 (.53) \\
Overview 			& 1 (N/A)& 4 (0)  &10 (.20)&10 (.13)&10 (.10)& 8 (.05)& 6 (.31)& 8 (.05)& 3 (0)  & 3 (.15)& 5 (.22)& 1 (.33)& 2 (.12)& 1 (0)  & 1 (.06)& \\
Media Type 			&        & 1 (1)  & 6 (.58)&8 (.38) & 8 (.83)& 4 (.35)& 3 (.33)& 7 (.41)& 2 (.19)& 4 (.28)& 8 (.28)& 1 (.50)&        & 2 (.36)&        & 1 (.16) \\
Author / Editor 		&        & 2 (1)  & 3 (.67)&2 (.67) & 4 (.62)&        & 3 (.61)& 6 (.50)& 4 (.50)&        & 4 (.68)& 1 (.75)& 7 (.34)& 3 (.83)&        & 4 (.53) \\
Content Types 			&        & 1 (1)  & 3 (.67)& 4 (.83)& 8 (.47)& 1 (1)  & 1 (1)  & 3 (.76)& 2 (.50)& 8 (.54)& 7 (.70)& 1 (.83)&        &        &        & 2 (.45) \\
Subjective Assessment 		& 1 (N/A)& 2 (1)  & 2 (.50)& 2 (.67)& 6 (.71)& 4 (.65)& 3 (.67)& 2 (1)  & 3 (.62)& 6 (.78)& 2 (.65)&        & 2 (.27)&        &        & \\
Topic 				&        & 4 (.50)& 2 (1)  & 2 (.67)& 8 (.28)& 2 (.30)& 1 (.67)& 4 (.57)&        & 5 (.36)& 3 (.27)&        & 3 (.44)&        &        & \\
Authority / Authoritativeness 	&  	 & 2 (.50)&        & 1 (.33)& 4 (.94)& 3 (.47)& 3 (.50)& 4 (.64)& 3 (.62)& 1 (.67)&        & 1 (0)  &        & 1 (.07)& 1 (0)  & 2 (.47) \\
Background / Source	 	&        &        & 2 (0)  & 4 (.33)& 2 (.38)& 1 (.20)&        & 2 (.21)& 1 (.38)& 1 (0)  & 3 (.13)& 2 (.12)& 2 (.88)&        &        & 1 (.68) \\
Navigation / Internal Structure & 	 &        &        &        & 1 (.75)&        & 2 (.50)&        & 1 (.88)& 5 (.56)& 2 (.55)& 2 (.33)&        & 1 (.50)&        & \\
Collection Size	 		&        & 1 (0)  & 1 (0)  & 2 (.83)& 2 (.38)&        & 1 (.17)& 1 (.57)&        & 1 (.22)& 2 (.60)&        &        &        &        & 2 (.24) \\
Purpose 			&        &        & 3 (.83)& 2 (.33)& 1 (.50)&        & 1 (.50)& 1 (.29)&        &        & 1 (.60)& 1 (1)  & 3 (.36)&        &        & \\
Audience 			&        & 1 (0)  &        & 3 (.33)& 3 (.42)&        &        & 2 (.79)& 1 (.62)&        &        & 1 (.92)& 1 (1)  &        &        & \\
Contributor 			&        &        &        & 3 (1)  & 2 (1)  & 2 (1)  &        & 1 (1)  & 1 (1)  &        & 1 (1)  &        & 1 (1)  &        &        & 1 (1) \\
Cross-resource Comparison	& 2 (N/A)&        & 1 (1)  & 2 (.33)&        & 3 (.60)&        &        &        &        & 3 (.50)&        &        &        &        & \\
Size/Length 			&        &        &        & 1 (0)  &        & 2 (.20)& 1 (.67)&        &        & 3 (.22)&        & 2 (.62)&        &        &        & \\
Style 				&        &        &        &        &        & 1 (.40)& 1 (.83)& 2 (.36)&        & 2 (.39)& 2 (.85)&        &        &        &        & \\
Query Relevance		 	&        &        &        &        &        &        &        &        & 2 (.75)&        &        &        &        &        &        & \\
Readability 			&        &        &        &        &        & 3 (.53)&        &        &        &        &        &        & 1 (.92)&        &        & \\
Difficulty 			&        &        &        & 3 (.67)&        &        &        &        &        & 1 (1)  &        &        &        &        &        & \\
Edition / Publication 		&        & 1 (0)  &        & 1 (0)  &        &        &        &        &        & 1 (1)  &        &        &        &        &        & \\
Language	 		&        & 1 (1)  & 1 (.50)&        &        &        &        &        &        &        &        &        &        &        &        & \\
Copyright 			&        &        &        &        &        &        &        &        &        & 2 (.94)&        &        &        &        &        & \\
Award 				&        &        &        &        &        & 2 (.70)&        &        &        &        &        &        &        &        &        & \\
\hline
\end{tabular}
\caption{Feature distribution across entries of different document
lengths.  Frequency of document feature given as entry, average
relative position of feature given in parentheses (0 indicates the
beginning of the entry, 1, the end of the entry).  Document features
listed in order of descending frequency in the annotated corpus.}
\label{t:featureDistribution}
\end{table*}

\section{Corpus miscellanea}

Command-line utilities also provided to modify, insert and extract
attributes from the corpus.  The web-based CGI scripts used by the
authors to build and analyze the corpus are also provided.

The corpus will be made web-accessible to licensed parties.  We would
like to encourage other research groups to join in expanding the
collection and annotation of additional bibliographic entries.

\subsection{Availability and copyright issues}

The corpus is available for academic and not-for-profit research, by
request to the first author.  A licensing agreement is required in
order to acquire the corpus and is available on the Columbia Natural
Language Group's ``Tools''
page\footnote{http://www.cs.columbia.edu/nlp/tools.html}.  An
annotation guide, explaining the annotation tagging guidelines in more
detail, will also be made available.

As the bibliographic entries themselves are mostly copyrighted by the
individual parties that have authored the entries, we can only
distribute the entries under the United States' Fair Use copyright
exemption, which allows the copying or excerpting of copyrighted text
for non-profit research and scholarship purposes.  Other for-profit
institutions interested in acquiring the corpus should also contact
the first author for information.  The delimitation and annotations of
the entries can be separated from the entry texts themselves using
standoff annotations and can be distributed; institutions can then
follow up with individual authors for rights to the source texts.

\section{Future work}

The corpus serves as a basis for our current research in
corpus-trained natural language generation.  In a high-level strategic
component, we establish ordering preferences between the document
features to determine when in the summary they occur.  In a low-level
tactical component, we find constraints on the lexical realization and
phrasing of the document features.  We are also in the continuing
process of refining our tagset (particularly in further
differentiating {\it detail} sentences into particular subclasses) and
collecting and annotating additional corpus entries.

\section{Conclusions}

We have presented our motivations for collecting a corpus of annotated
bibliography entries, as a means of studying appropriate summary forms
for documents in information retrieval displays.  Annotated
bibliography entries are constructed by abstractive techniques and
display both indicative and informative qualities.  While topical,
content based features are prominent and necessary in summaries,
guidelines have suggested that summaries should also include metadata
and critical document features.  Our corpus study has shown that these
guidelines are followed in actual annotated entries, and furthermore
have quantitatively assessed their importance and explored their
internal ordering within summaries of different lengths.

We have detailed the methodology used to collect the 2000-entry corpus
and detailed our annotation and document feature distribution across
100 randomly selected entries.  The corpus is available for non-profit
research use and we would like to encourage other researchers to use
and contribute to this corpus as well.

\section{Acknowledgments}

This research is supported by the National Science Foundation under
Digital Library Initiative Phase II Grant Number IIS-98-17434.  We
would also like to acknowledge the Linguistic Data Consortium and our
local legal consel in helping us clarify intellectual property issues
involved with this work.


\bibliographystyle{lrec2000}
\bibliography{lrec02} 

\end{document}